\documentclass[runningheads]{llncs}
\usepackage{graphicx}
\usepackage{amsmath,amssymb} 
\usepackage{color}
\usepackage{breakcites}
\usepackage{hyperref}

\begin{document}
\title{Stacked Cross Attention for \newline Image-Text Matching} 

\titlerunning{Stacked Cross Attention for Image-Text Matching}
%
\author{Kuang-Huei Lee\inst{1} \and Xi Chen\inst{1} \and Gang Hua\inst{1} \and Houdong Hu\inst{1} \and Xiaodong He\inst{2}\thanks{Work performed while working at Microsoft Research.}}
%
\authorrunning{Kuang-Huei Lee, Xi Chen, Gang Hua, Houdong Hu, Xiaodong He}
%

\institute{Microsoft AI and Research \\
\email{\{kualee,chnxi,ganghua,houhu\}@microsoft.com}\and
JD AI Research\\
\email{xiaodong.he@jd.com}}
\maketitle              
\begin{abstract}
In this paper, we study the problem of image-text matching. Inferring the latent semantic alignment between objects or other salient stuff ({\em e.g.} snow, sky, lawn) and the corresponding words in sentences allows to capture fine-grained interplay between vision and language, and makes image-text matching more interpretable. Prior work either simply aggregates the similarity of all possible pairs of regions and words without attending differentially to more and less important words or regions, or uses a multi-step attentional process to capture limited number of semantic alignments which is less interpretable. In this paper, we present \textit{Stacked Cross Attention} to discover the full latent alignments using both image regions and words in a sentence as context and infer image-text similarity. Our approach achieves the state-of-the-art results on the MS-COCO and Flickr30K datasets. On Flickr30K, our approach outperforms the current best methods by 22.1\% relatively in text retrieval from image query, and 18.2\% relatively in image retrieval with text query (based on Recall@1). On MS-COCO, our approach improves sentence retrieval by 17.8\% relatively and image retrieval by 16.6\% relatively (based on Recall@1 using the 5K test set). Code has been made available at: \url{https://github.com/kuanghuei/SCAN}.

\end{abstract}
\section{Introduction}

In this paper we study the problem of image-text matching, central to image-sentence cross-modal retrieval ({\em i.e.} image search for given sentences with visual descriptions and the retrieval of sentences from image queries). 

\begin{figure}[t!]
\centering
\includegraphics[height=5.0cm]{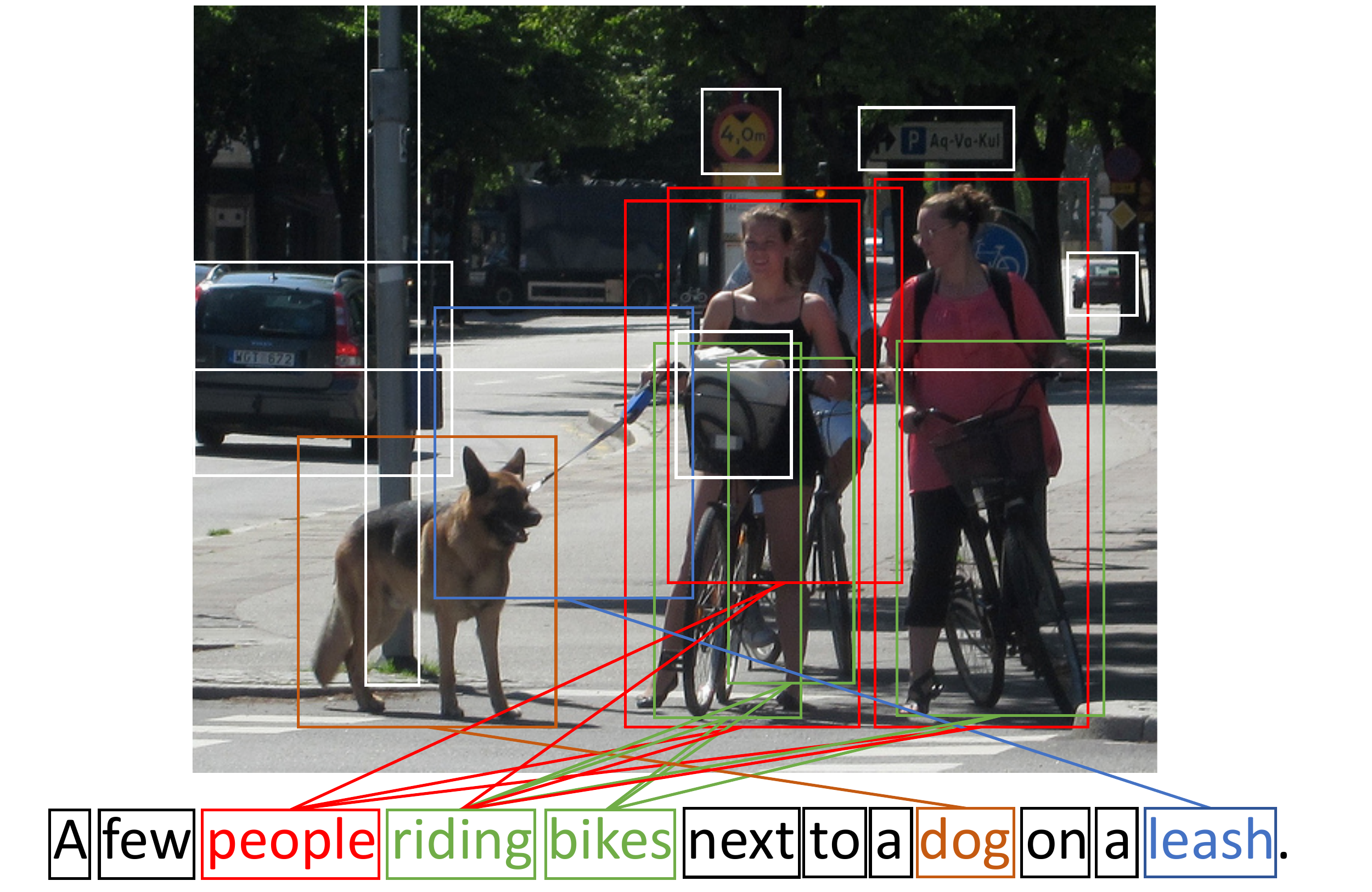}
\caption{Sentence descriptions make frequent reference to some particular but unknown salient regions in images, as well as their attributes and actions. Reasoning the underlying correspondence is a key to interpretable image-text matching.}
\label{fig:concept}
\end{figure}

When people describe what they see, it can be observed that the descriptions make frequent reference to objects and other salient stuff in the images, as well as their attributes and actions (as shown in Figure~\ref{fig:concept}). In a sense, sentence descriptions are weak annotations, where words in a sentence correspond to some particular, but unknown regions in the image. Inferring the latent correspondence between image regions and words is a key to more interpretable image-text matching by capturing the fine-grained interplay between vision and language.

Similar observations motivated prior work on image-text matching~\cite{karpathy2015deep,karpathy2014deep,niu2017hierarchical}. These models often detect image regions at object/stuff level and simply aggregate the similarity of all possible pairs of image regions and words in sentence to infer the global image-text similarity; {\em e.g.} Karpathy and Fei-Fei~\cite{karpathy2015deep} proposed taking the maximum of the region-word similarity scores with respect to each word and averaging the results corresponding to all words. It shows the effectiveness of inferring the latent region-word correspondences, but such aggregation does not consider the fact that the importance of words can depend on the visual context.  

We strive to take a step towards attending differentially to important image regions and words with each other as context for inferring the image-text similarity. We introduce a novel \textit{Stacked Cross Attention} that enables attention with context from both image and sentence in two stages. In the proposed \textit{Image-Text} formulation, given an image and a sentence, it first attends to words in the sentence with respect to each image region, and compares each image region to the attended information from the sentence to decide the importance of the image regions ({\em e.g.} mentioned in the sentence or not). Likewise, in the proposed \textit{Text-Image} formulation, it first attends to image regions with respect to each word and then decides to pay more or less attention to each word. 

Compared to models that perform fixed-step attentional reasoning and thus only focus on limited semantic alignments (one at a time) \cite{nam2016dual,huang2017instance}, Stacked Cross Attention discovers all possible alignments simultaneously. Since the number of semantic alignments varies with different images and sentences, the correspondence inferred by our method is more comprehensive and thus making image-text matching more interpretable. 

To identify the salient regions in image, we follow Anderson \textit{et al.} \cite{anderson2017bottom} to analogize the detection of salient regions at object/stuff level to the spontaneous bottom-up attention in the human vision system \cite{buschman2007top,corbetta2002control,katsuki2014bottom}, and practically implement bottom-up attention using Faster R-CNN \cite{ren2015faster}, which represents a natural expression of a bottom-up attention mechanism. 

To summarize, our primary contribution is the novel Stacked Cross Attention mechanism for discovering the full latent visual-semantic alignments. To evaluate the performance of our approach in comparison to other architectures and perform comprehensive ablation studies, we look at the MS-COCO \cite{lin2014microsoft} and Flickr30K \cite{young2014image} datasets. Our model, Stacked Cross Attention Network (SCAN) that uses the proposed attention mechanism, achieves the state-of-the-art results. On Flickr30K, our approach outperforms the current best methods by 22.1\% relatively in text retrivel from image query, and 18.2\% relatively in image retrieval with text query (based on Recall@1). On MS-COCO, it improves sentence retrieval by 17.8\% relatively and image retrieval by 16.6\% relatively (based on Recall@1 using the 5K test set). 

\section{Related Work}

A rich line of studies have explored mapping whole images and full sentences to a common semantic vector space for image-text matching \cite{kiros2014unifying,vendrov2015order,ba2016layer,wang2016learning,klein2015associating,lev2016rnn,zheng2017dual,faghri2017vse++,peng2017cm,gu2017look,eisenschtat2017linking,devlin2015language,fang2015captions}. Kiros \textit{et al.} \cite{kiros2014unifying} made the first attempt to learn cross-view representations with a hinge-based triplet ranking loss using deep Convolutional Neural Networks (CNN) to encode images and Recurrent Neural Networks (RNN) to encode sentences. Faghri \textit{et al.} \cite{faghri2017vse++} leveraged hard negatives in the triplet loss function and yielded significant improvement. Peng \textit{et al.} \cite{peng2017cm} and Gu \textit{et al.} \cite{gu2017look} suggested incorporating generative objectives into the cross-view feature embedding learning. As opposed to our proposed method, the above works do not consider the latent vision-language correspondence at the level of image regions and words. Specifically, we discuss two lines of research addressing this problem using attention mechanism as follows.

\noindent
\textbf{Image-text matching with bottom-up attention. } Bottom-up attention is a terminology that Anderson \textit{et al.} \cite{anderson2017bottom} proposed in their work on image captioning and Visual Question-Answering (VQA), referring to purely visual feed-forward attention mechanisms in analogy to the spontaneous bottom-up attention in human vision system \cite{buschman2007top,corbetta2002control,katsuki2014bottom} ({\em e.g.} human attention tends to be attracted to salient instances like objects instead of background). Similar observation had motivated this study and several other works \cite{karpathy2015deep,karpathy2014deep,niu2017hierarchical,huang2017learning}. Karpathy and Fei-Fei \cite{karpathy2015deep} proposed detecting and encoding image regions at object level with R-CNN \cite{girshick2014rich}, and then inferring the image-text similarity by aggregating the similarity scores of all possible region-word pairs. Niu \textit{et al.} \cite{niu2017hierarchical} presented a model that maps noun phrases within sentences and objects in images into a shared embedding space on top of full sentences and whole images embeddings. Huang \textit{et al.} \cite{huang2017learning} combined image-text matching and sentence generation for model learning with an improved image representation including objects, properties, actions, etc. In contrast to our model, these studies do not use the conventional attention mechanism ({\em e.g.} \cite{xu2015show}) to learn to focus on image regions for given semantic context.

\noindent
\textbf{Conventional attention-based methods. } The attention mechanism focuses on certain aspects of data with respect to a task-specific context ({\em e.g.} looking for something). In computer vision, visual attention aims to focus on specific images or subregions \cite{anderson2017bottom,xu2015show,xu2017attngan,lee2017cleannet}. Similarly, attention methods for natural language processing adaptively select and aggregate informative snippets to infer results \cite{yang2016hierarchical,rush2015neural,bahdanau2014neural,kumar2016ask,li2015hierarchical}. Recently, attention-based models have been proposed for the image-text matching problem. Huang \textit{et al.} \cite{huang2017instance} developed a context-modulated attention scheme to selectively attend to a pair of instances appearing in both the image and sentence. Similarly, Nam \textit{et al.} \cite{nam2016dual} proposed Dual Attentional Network to capture fine-grained interplay between vision and language through multiple steps. However, these models adopt multi-step reasoning with a pre-defined number of steps to look at one semantic matching ({\em e.g.} an object in the image and a phrase in the sentence) at a time, despite the number of semantic matchings change for different images and sentence descriptions. In contrast, our proposed model discovers all latent alignments, thus is more interpretable.

\section{Learning Alignments with Stacked Cross Attention}

In this section, we describe the Stacked Cross Attention Network (SCAN). Our objective is to map words and image regions into a common embedding space to infer the similarity between a whole image and a full sentence. We begin with bottom-up attention to detect and encode image regions into features. Also, we map words in sentence along with the sentence context to features. We then apply Stacked Cross Attention to infer the image-sentence similarity by aligning image region and word features. We first introduce Stacked Cross Attention in Section \ref{sec:sq_attn} and the objective of learning alignments in Section \ref{sec:ali_obj}. Then we detail image and sentence representations in Section \ref{sec:image_rep} and Section \ref{sec:text_rep}, respectively.

\subsection{Stacked Cross Attention}
\label{sec:sq_attn}
Stacked Cross Attention expects two inputs: a set of image features $V = \{v_1, ..., v_k\allowbreak\}, v_i \in \mathbb{R}^D$, such that each image feature encodes a region in an image; a set of word features $E = \{e_1, ..., e_n\}, e_i \in \mathbb{R}^D$, in which each word feature encodes a word in a sentence. The output is a similarity score, which measures the similarity of an image-sentence pair. In a nutshell, Stacked Cross Attention attends differentially to image regions and words using both as context to each other while inferring the similarity. We define two complimentary formulations of Stacked Cross Attention below: \textit{Image-Text} and \textit{Text-Image}.

\begin{figure}[t]
\centering
\includegraphics[width=\textwidth]{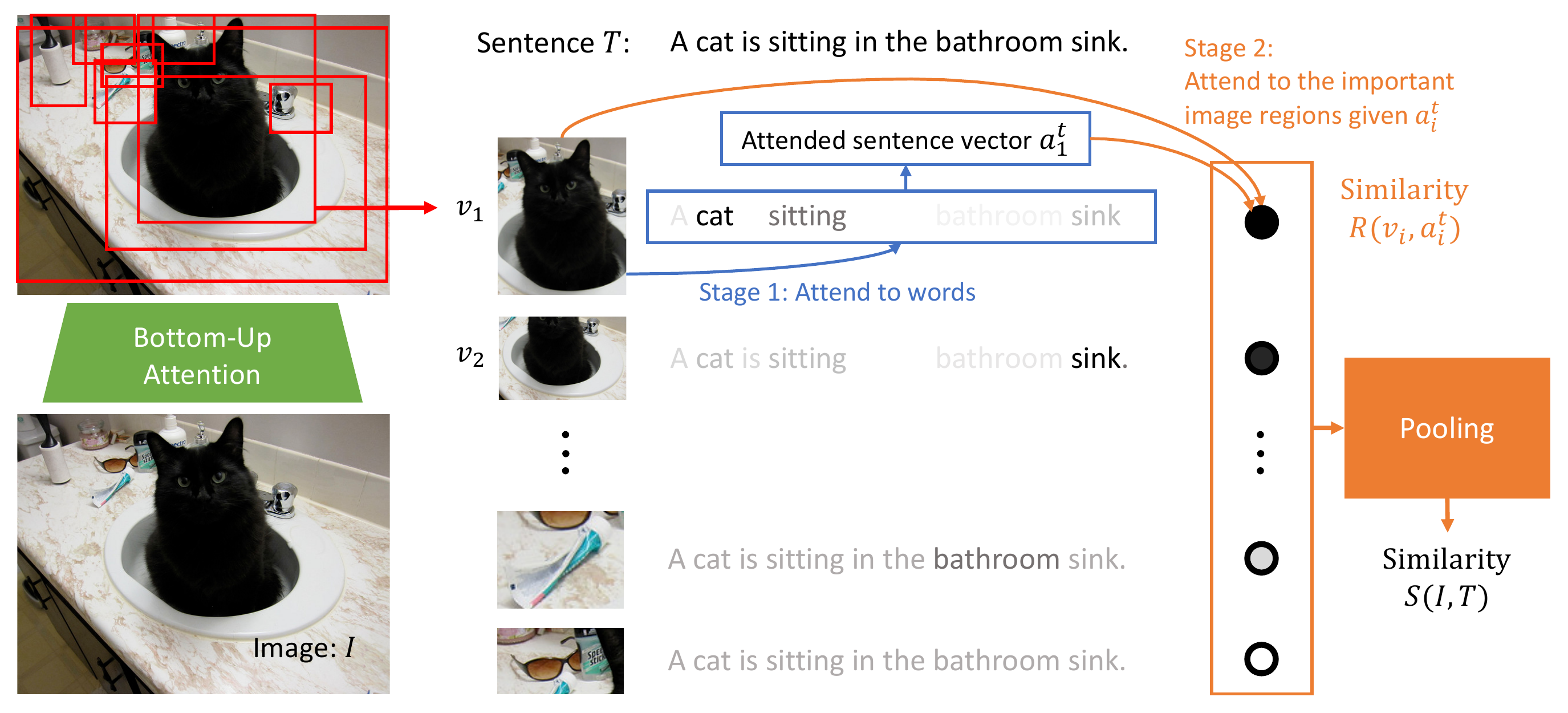}
\caption{Image-Text Stacked Cross Attention: At stage 1, we first attend to words in the sentence with respect to each image region feature $v_i$ to generate an attended sentence vector $a_i^t$ for $i$-th image region. At stage 2, we compare $a_i^t$ and $v_i$ to determine the importance of each image region, and then compute the similarity score.}
\label{fig:sqattn_i2t}
\end{figure}

\noindent
\textbf{Image-Text Stacked Cross Attention.} This formulation is illustrated in Figure \ref{fig:sqattn_i2t}, entailing two stages of attention. First, it attends to words in the sentence with respect to each image region. In the second stage, it compares each image region to the corresponding attended sentence vector in order to determine the importance of the image regions with respect to the sentence. Specifically, given an image $I$ with $k$ detected regions and a sentence $T$ with $n$ words, we first compute the cosine similarity matrix for all possible pairs, {\em i.e.}
\begin{equation}
\label{eq:cosine_sim}
s_{ij} = \dfrac{v_i^Te_j}{||v_i||||e_j||}, i \in [1, k], j \in [1, n].
\end{equation}
Here, $s_{ij}$ represents the similarity between the $i$-th region and the $j$-th word. We empirically find it beneficial to threshold the similarities at zero \cite{karpathy2014deep} and normalize the similarity matrix as $\bar{s}_{ij} = [s_{ij}]_+/\sqrt[]{\sum_{i=1}^k [s_{ij}]_+^2}$, where $[x]_+ \equiv max(x,0)$.

To attend on words with respect to each image region, we define a weighted combination of word representations ({\em i.e.} the attended sentence vector $a_i^t$, with respect to the $i$-th image region)

\begin{equation}
a_i^t = \sum_{j=1}^n\alpha_{ij}e_j,
\end{equation}
where
\begin{equation}
\alpha_{ij}= \dfrac{exp(\lambda_1\bar{s}_{ij})}{\sum_{j=1}^n exp(\lambda_1\bar{s}_{ij})},
\label{eq:softmax_i2t}
\end{equation}
and $\lambda_1$ is the inversed temperature of the softmax function \cite{chorowski2015attention} (Eq.~\eqref{eq:softmax_i2t}). This definition of attention weights is a variant of dot product attention \cite{luong2015effective}.

To determine the importance of each image region given the sentence context, we define relevance between the $i$-th region and the sentence as cosine similarity between the attended sentence vector $a_i^t$ and each image region feature $v_i$, {\em i.e.}
\begin{equation}
R(v_i,a_i^t) = \dfrac{v_i^Ta_i^t}{||v_i||||a_i^t||}.
\end{equation}
Inspired by the minimum classification error formulation in speech recognition \cite{juang1997minimum,he2008discriminative}, the similarity between image $I$ and sentence $T$ is calculated by LogSumExp pooling (LSE), {\em i.e.}
\begin{equation}
S_{LSE}(I, T) = log(\sum_{i=1}^{k} exp(\lambda_2 R(v_i,a_i^t)))^{(1/\lambda_2)},
\end{equation}
where $\lambda_2$ is a factor that determines how much to magnify the importance of the most relevant pairs of image region feature $v_i$ and attended sentence vector $a_i^t$. As $\lambda_2 \rightarrow \infty$, $S(I, T)$ approximates to $max_{i=1}^{k}R(v_i,a_i^t)$.
Alternatively, we can summarize $R(v_i,a_i^t)$ with average pooling (AVG), {\em i.e.}
\begin{equation}
S_{AVG}(I, T) = \dfrac{\sum_{i=1}^{k} R(v_i,a_i^t)}{k}.
\end{equation}
Essentially, if region $i$ is not mentioned in the sentence, its feature $v_i$ would not be similar to the corresponding attended sentence vector $a_i^t$ since it would not be able to collect good information while computing $a_i^t$. Thus, comparing $a_i^t$ and $v_i$ determines how important region $i$ is with respect to the sentence.

\begin{figure}[t]
\centering
\includegraphics[width=\textwidth]{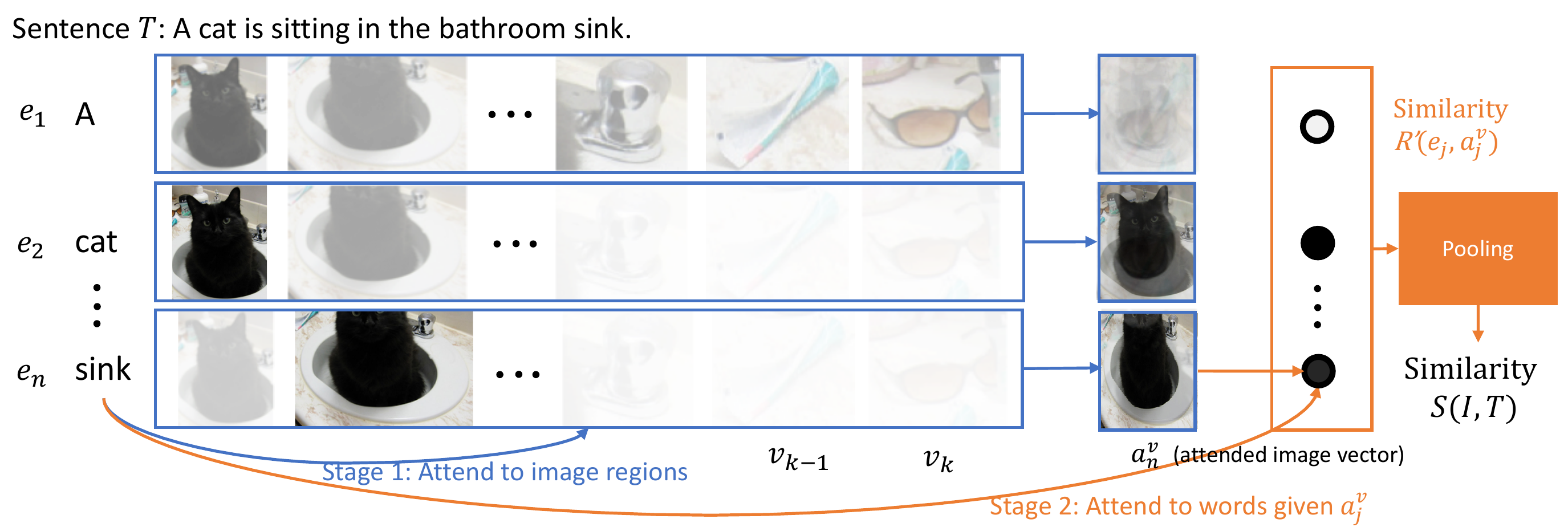}
\caption{Text-Image Stacked Cross Attention:  At stage 1, we first attend to image regions with respect to each word feature $e_i$ to generate an attended image vector $a_j^v$ for $j$-th word in the sentence (The images above the symbol $a_n^v$ represent the attended image vectors). At stage 2, we compare $a_j^v$ and $e_j$ to determine the importance of each image region, and then compute the similarity score.}
\label{fig:sqattn_t2i}
\end{figure}

\noindent
\textbf{Text-Image Stacked Cross Attention}. Likewise, we can first attend to image regions with respect to each word, and compare each word to the corresponding attended image vector to determine the importance of each word. We call this formulation \textit{Text-Image}, which is depicted in Figure~\ref{fig:sqattn_t2i}. Specifically, we normalize cosine similarity $s_{i,j}$ between the $i$-th region and the $j$-th word as $\bar{s}'_{i,j} = [s_{i,j}]_+/\sqrt[]{\sum_{j=1}^n [s_{i,j}]_+^2}$.

To attend on image regions with respect to each word, we define a weighted combination of image region features ({\em i.e.} the attended image vector $a_j^v$ with respect to $j$-th word): $a_j^v = \sum_{i=1}^k\alpha_{ij}'v_i$, where $\alpha_{ij}'= exp(\lambda_1\bar{s}'_{i,j})/\sum_{i=1}^k exp(\lambda_1\bar{s}'_{i,j})$.
Using the cosine similarity between the attended image vector $a_j^v$ and the word feature $e_j$, we measure the relevance between the $j$-th word and the image as $R'(e_j,a_j^v) = (e_j^Ta_j^v)/(||e_j||||a_j^v||)$. The final similarity score between image $I$ and sentence $T$ is summarized by LogSumExp pooling (LSE), {\em i.e.}
\begin{equation}
S_{LSE}'(I, T) = log(\sum_{j=1}^{n} exp(\lambda_2 R'(e_j,a_j^v)))^{(1/\lambda_2)},
\end{equation}
or alternatively by average pooling (AVG)
\begin{equation}
S_{AVG}'(I, T) = \dfrac{\sum_{j=1}^{n} R'(e_j,a_j^v)}{n}. 
\end{equation}
In prior work, Karpathy and Fei-Fei \cite{karpathy2015deep} defined region-word similarity as a dot product between $v_i$ and $e_j$, {\em i.e.} $s_{ij} = v_i^Te_j$ and image-text similarity by aggregating all possible pairs without attention as
\begin{equation}
S_{SM}'(I,T) = \sum_{j=1}^n\max_{i}(s_{ij}).
\label{eq:sum_max_ti}
\end{equation}
We revisit this formulation in our ablation studies in Section~\ref{sec:ablation}, dubbed \textit{Sum-Max Text-Image}, and also the symmetric form, dubbed \textit{Sum-Max Image-Text}
\begin{equation}
S_{SM}(I,T) = \sum_{i=1}^k\max_{j}(s_{ij}).
\label{eq:sum_max_it}
\end{equation}

\subsection{Alignment Objective}
\label{sec:ali_obj}
Triplet loss is a common ranking objective for image-text matching. Previous approaches~\cite{karpathy2015deep,kiros2014unifying,socher2014grounded} have employed a hinge-based triplet ranking loss with margin $\alpha$, {\em i.e.}
\begin{equation}
l(I, T) = \sum_{\hat{T}} [\alpha - S(I,T) + S(I,\hat{T})]_{+} + \sum_{\hat{I}} [\alpha - S(I,T) + S(\hat{I},T)]_+,
\label{triplet_loss}
\end{equation}
where $[x]_+ \equiv max(x,0)$ and $S$ is a similarity score function (\textit{e.g.} $S_{LSE}$). The first sum is taken over all negative sentences $\hat{T}$ given an image $I$; the second sum considers all negative images $\hat{I}$ given a sentence $T$. If $I$ and $T$ are closer to one another in the joint embedding space than to any negatives pairs, by the margin $\alpha$, the hinge loss is zero. In practice, for computational efficiency, rather than summing over all the negative samples, it usually considers only the hard negatives in a mini-batch of stochastic gradient descent.

In this study, we focus on the hardest negatives in a mini-batch following Fagphri~\textit{et al.}~\cite{faghri2017vse++}. For a positive pair $(I, T)$, the hardest negatives are given by $\hat{I}_h = argmax_{m\neq I}S(m,T)$ and $\hat{T}_h = argmax_{d\neq T}S(I,d)$. We therefore define our triplet loss as
\begin{equation}
l_{hard}(I, T) = [\alpha - S(I,T) + S(I,\hat{T}_h)]_{+} + [\alpha - S(I,T) + S(\hat{I}_h,T)]_+.
\label{triplet_loss_hard}
\end{equation}

\subsection{Representing images with Bottom-Up Attention}
\label{sec:image_rep}
Given an image $I$, we aim to represent it with a set of image features $V = \{v_1, ..., v_k\}, v_i \in \mathbb{R}^D$, such that each image feature encodes a region in an image. The definition of an image region is generic. However, in this study, we focus on regions at the level of object and other entities. Following Anderson~\textit{et al.}~\cite{anderson2017bottom}. we refer to detection of salient regions as bottom-up attention and practically implement it with a Faster R-CNN~\cite{ren2015faster} model. 

Faster R-CNN is a two-stage object detection framework. In the first stage of Region Proposal Network (RPN), a grid of anchors tiled in space, scale and aspect ratio are used to generate bounding boxes, or Region Of Interests (ROIs), with high objectness scores. In the second stage the representations of the ROIs are pooled from the intermediate convolution feature map for region-wise classification and bounding box regression. A multi-task loss considering both classification and localization are minimized in both the RPN and final stages.

We adopt the Faster R-CNN model in conjunction with ResNet-101~\cite{he2016deep} pre-trained by Anderson~\textit{et al.}~\cite{anderson2017bottom} on Visual Genomes~\cite{krishna2017visual}. In order to learn feature representations with rich semantic meaning, instead of predicting the object classes, the model predicts attribute classes and instance classes, in which instance classes contain objects and other salient stuff that is difficult to localize ({\em e.g.} stuff like `sky', `grass', `building' and attributes like `furry'). 

For each selected region $i$, $f_i$ is defined as the mean-pooled convolutional feature from this region, such that the dimension of the image feature vector is 2048. We add a fully-connect layer to transform $f_i$ to a $h$-dimensional vector
\begin{equation}
v_i = W_vf_i + b_v.
\end{equation}
Therefore, the complete representation of an image is a set of embedding vectors $v = \{v_1, ..., v_k\}, v_i \in \mathbb{R}^D$, where each $v_i$ encodes an salient region and $k$ is the number of regions. 

\subsection{Representing Sentences}
\label{sec:text_rep}

To connect the domains of vision and language, we would like to map language to the same $h$-dimensional semantic vector space as image regions. Given a sentence $T$, the simplest approach is mapping every word in it individually. However, this approach does not consider any semantic context in the sentence. Therefore, we employ an RNN to embed the words along with their context.

For the $i$-th word in the sentence, we represent it with an one-hot vector showing the index of the word in the vocabulary, and embed the word into a 300-dimensional vector through an embedding matrix $W_e$. $x_i = W_e w_i, i \in [1, n]$. We then use a bi-directional GRU~\cite{bahdanau2014neural,schuster1997bidirectional} to map the vector to the final word feature along with the sentence context by summarizing information from both directions in the sentence. The bi-directional GRU contains a forward GRU which reads the sentence $T$ from $w_1$ to $w_n$
\begin{equation}
\overrightarrow{h_i} = \overrightarrow{GRU}(x_i), i \in [1, n]
\end{equation}
and a backward GRU which reads from $w_n$ to $w_1$
\begin{equation}
\overleftarrow{h_i} = \overleftarrow{GRU}(x_i), i \in [1, n].
\end{equation}
The final word feature $e_i$ is defined by averaging the forward hidden state $\overrightarrow{h_i}$ and backward hidden state $\overleftarrow{h_i}$, which summarizes information of the sentence centered around $w_i$
\begin{equation}
e_i = \dfrac{(\overrightarrow{h_i} + \overleftarrow{h_i})}{2}, i \in [1, n].
\end{equation}

\section{Experiments}
We carry out extensive experiments to evaluate Stacked Cross Attention Network (SCAN), and compare various formulations of SCAN to other state-of-the-art approaches. We also conduct ablation studies to incrementally verify our approach and thoroughly investigate the behavior of SCAN. As is common in information retreival, we measure performance of sentence retrieval (image query) and image retrieval (sentence query) by recall at $K$ (R@$K$) defined as the fraction of queries for which the correct item is retrieved in the closest $K$ points to the query. The hyperparameters of SCAN, such as $\lambda_1$ and $\lambda_2$, are selected on the validation set. Details of training and the bottom-up attention implementation are presented in the supplementary material.

\subsection{Datasets}
We evaluate our approach on the MS-COCO and Flickr30K datasets. Flickr30K contains 31,000 images collected from Flickr website with five captions each. Following the split in \cite{karpathy2015deep,faghri2017vse++}, we use 1,000 images for validation and 1,000 images for testing and the rest for training. MS-COCO contains 123,287 images, and each image is annotated with five text descriptions. In \cite{karpathy2015deep}, the dataset is split into 82,783 training images, 5,000 validation images and 5,000 test images. We follow \cite{faghri2017vse++} to add 30,504 images that were originally in the validation set of MS-COCO but have been left out in this split into the training set. Each image comes with 5 captions. The results are reported by either averaging over 5 folds of 1K test images or testing on the full 5K test images. Note that some early works such as \cite{karpathy2015deep} only use a training set containing 82,783 images.

\subsection{Results on Flickr30K}
\label{sec:res_f30k}

\begin{table}[t!]
\begin{center}
\caption{Comparison of the cross-modal retrieval restuls in terms of Recall@$K$(R@$K$) on Flickr30K. t-i denotes Text-Image. i-t denotes Image-Text. AVG and LSE denotes average and LogSumExp pooling respectively.}
\label{table:flickr30k}
\begin{tabular}
{p{5.0cm}p{1.0cm}p{1.0cm}p{1.0cm}p{1.0cm}p{1.0cm}p{1.0cm}}
\hline\noalign{\smallskip}
 & \multicolumn{3}{c}{Sentence Retrieval} & \multicolumn{3}{c}{Image Retrieval} \\
Method & R@1 & R@5 & R@10 & R@1 & R@5 & R@10 \\
\noalign{\smallskip}
\hline
\noalign{\smallskip}
DVSA (R-CNN, AlexNet) \cite{karpathy2015deep} & 22.2 & 48.2 & 61.4 & 15.2 & 37.7 & 50.5 \\ 
HM-LSTM (R-CNN, AlexNet) \cite{niu2017hierarchical} & 38.1 & - & 76.5 & 27.7 & - & 68.8 \\ 
SM-LSTM (VGG) \cite{huang2017instance} & 42.5 & 71.9 & 81.5 & 30.2 & 60.4 & 72.3 \\ 
2WayNet (VGG) \cite{eisenschtat2017linking} & 49.8 & 67.5 & - & 36.0 & 55.6 & - \\ 
DAN (ResNet) \cite{nam2016dual} & 55.0 & 81.8 & 89.0 & 39.4 & 69.2 & 79.1 \\ 
VSE++ (ResNet) \cite{faghri2017vse++} & 52.9 & - & 87.2 & 39.6 & - & 79.5 \\ 
DPC (ResNet) \cite{zheng2017dual} & 55.6 & 81.9 & 89.5 & 39.1 & 69.2 & 80.9 \\ 
SCO (ResNet) \cite{huang2017learning} & 55.5 & 82.0 & 89.3 & 41.1 & 70.5 & 80.1 \\ 
\hline
Ours (Faster R-CNN, ResNet):  \\ 
SCAN t-i LSE ($\lambda_1=9,\lambda_2=6$)  & 61.1 & 85.4 & 91.5 & 43.3 & 71.9 & 80.9  \\ 
SCAN t-i AVG ($\lambda_1=9$)  & 61.8 & 87.5 & 93.7 & 45.8 & 74.4 & 83.0  \\ 
SCAN i-t LSE ($\lambda_1=4,\lambda_2=5$)   & 67.7 & 88.9 & 94.0 & 44.0 & 74.2 & 82.6 \\ 
 
SCAN i-t AVG ($\lambda_1=4$) & \textbf{67.9} & 89.0 & 94.4 & 43.9 & 74.2 & 82.8  \\ 
SCAN t-i AVG + i-t LSE  & 67.4  & \textbf{90.3} & \textbf{95.8} & \textbf{48.6} & \textbf{77.7} & \textbf{85.2} \\
\hline
\end{tabular}
\end{center}
\end{table}
\setlength{\tabcolsep}{1.4pt}

Table~\ref{table:flickr30k} presents the quantitative results on Flickr30K where all formulations of our proposed method outperform recent approaches in all measures. We denote the Text-Image formulation by t-i, Image-Text formulation by i-t, LogSumExp pooling by LSE, and average pooling by AVG. The best R@1 of sentence retrieval given an image query is 67.9, achieved by SCAN i-t AVG, where we see a 22.1\% relative improvement comparing to DPC \cite{zheng2017dual}. Furthermore, we combine t-i and i-t models by averaging their predicted similarity scores. 
The best result of model ensembles is achieved by combining t-i AVG and i-t LSE, selected on the validation set. The combined model gives 48.6 at R@1 for image retrieval, which is a 18.2\% relative improvement from the current state-of-the-art, SCO \cite{huang2017learning}. Our assumption is that different formulations of Stacked Cross Attention (t-i and i-t; AVG/LSE pooling) approach different aspects of data, such that the model ensemble further improves the results.

\subsection{Results on MS-COCO}
\label{sec:res_coco}

Table \ref{table:mscoco} lists the experimental results on MS-COCO and a comparison with prior work. On the 1K test set, the single SCAN t-i AVG achieves comparable results to the current state-of-the-art, SCO. Our best result on 1K test set is achieved by combining t-i LSE and i-t AVG which improves 4.0\% on image query and 8.0\% relatively comparing to SCO. On the 5K test set, we choose to list the best single model and ensemble selected on the validation set due to space limitation. Both models outperform SCO on all metrics, and SCAN t-i AVG + i-t LSE improves 17.8\% on sentence retrieval (R@1) and 16.6\% on image retrieval (R@1) relatively.

\begin{table}[t!]
\begin{center}
\caption{Comparison of the cross-modal retrieval restuls in terms of Recall@$K$(R@$K$) on MS-COCO. t-i denotes Text-Image. i-t denotes Image-Text. AVG and LSE denotes average and LogSumExp pooling respectively.}
\label{table:mscoco}
\begin{tabular}
{p{5.0cm}p{1.0cm}p{1.0cm}p{1.0cm}p{1.0cm}p{1.0cm}p{1.0cm}}
\hline\noalign{\smallskip}
 & \multicolumn{3}{c}{Sentence Retrieval} & \multicolumn{3}{c}{Image Retrieval} \\
Method & R@1 & R@5 & R@10 & R@1 & R@5 & R@10 \\
\noalign{\smallskip}
\hline
\noalign{\smallskip}
& \multicolumn{6}{c}{1K Test Images} \\ 
\hline
\hline
DVSA (R-CNN, AlexNet) \cite{karpathy2015deep} & 38.4 & 69.9 & 80.5 & 27.4 & 60.2 & 74.8 \\ 
HM-LSTM (R-CNN, AlexNet) \cite{niu2017hierarchical} & 43.9 & - & 87.8 & 36.1 & - & 86.7 \\ 
Order-embeddings (VGG) \cite{vendrov2015order} & 46.7 & - & 88.9 & 37.9 & - & 85.9 \\ 
SM-LSTM (VGG) \cite{huang2017instance} & 53.2 & 83.1 & 91.5 & 40.7 & 75.8 & 87.4 \\ 
2WayNet (VGG) \cite{eisenschtat2017linking} & 55.8 & 75.2 & - & 39.7 & 63.3 & - \\ 
VSE++ (ResNet) \cite{faghri2017vse++} & 64.6 & - & 95.7 & 52.0 & - & 92.0 \\ 
DPC (ResNet) \cite{zheng2017dual} & 65.6 & 89.8 & 95.5 & 47.1 & 79.9 & 90.0 \\ 
GXN (ResNet) \cite{gu2017look} & 68.5 & - & 97.9 & 56.6 & - & 94.5 \\ 
SCO (ResNet) \cite{huang2017learning} & 69.9 & 92.9 & 97.5 & 56.7 & 87.5 & \textbf{94.8} \\ 
\hline
Ours (Faster R-CNN, ResNet):  \\ 
SCAN t-i LSE ($\lambda_1=9,\lambda_2=6$)  & 67.5 & 92.9 & 97.6 & 53.0 & 85.4 & 92.9 \\ 
SCAN t-i AVG ($\lambda_1=9$) & 70.9 & 94.5 & 97.8 & 56.4 & 87.0 & 93.9  \\ 
SCAN i-t LSE ($\lambda_1=4,\lambda_2=20$) & 68.4 & 93.9 & 98.0 & 54.8 & 86.1 & 93.3 \\ 
SCAN i-t AVG ($\lambda_1=4$) & 69.2 & 93.2 & 97.5 & 54.4 & 86.0 & 93.6 \\ 
SCAN t-i LSE + i-t AVG & \textbf{72.7} & \textbf{94.8} & \textbf{98.4} & \textbf{58.8} & \textbf{88.4} & \textbf{94.8} \\ 
\hline
& \multicolumn{6}{c}{5K Test Images} \\ 
\hline
\hline
Order-embeddings (VGG) \cite{vendrov2015order} & 23.3 & - & 84.7 & 31.7 & - & 74.6 \\ 
VSE++ (ResNet) \cite{faghri2017vse++} & 41.3 & - & 81.2 & 30.3 & - & 72.4 \\ 
DPC (ResNet) \cite{zheng2017dual} & 41.2 & 70.5 & 81.1 & 25.3 & 53.4 & 66.4 \\ 
GXN (ResNet) \cite{gu2017look} & 42.0 & - & 84.7 & 31.7 & - & 74.6 \\ 
SCO (ResNet) \cite{huang2017learning} & 42.8 & 72.3 & 83.0 & 33.1 & 62.9 & 75.5 \\ 
\hline
Ours (Faster R-CNN, ResNet):  \\ 
SCAN i-t LSE & 46.4 & 77.4 & 87.2 & 34.4 & 63.7 & 75.7 \\ 
SCAN t-i AVG + i-t LSE & \textbf{50.4} & \textbf{82.2} & \textbf{90.0} & \textbf{38.6} & \textbf{69.3} & \textbf{80.4} \\ 
\hline
\end{tabular}
\end{center}
\end{table}
\setlength{\tabcolsep}{1.4pt}

\subsection{Ablation Studies}
\label{sec:ablation}

\begin{table}[t!]
\begin{center}
\caption{Effect of inferring the latent vision-language alignment at the level of regions and words. Results are reported in terms of Recall@$K$(R@$K$). Refer to Eqs. \eqref{eq:sum_max_ti} \eqref{eq:sum_max_it} for the definition of Sum-Max. t-i denotes Text-Image. i-t denotes Image-Text.} 
\label{table:eff_sqattn}
\begin{tabular}
{p{5.0cm}p{1.0cm}p{1.0cm}p{1.0cm}p{1.0cm}p{1.0cm}p{1.0cm}}
\hline\noalign{\smallskip}
 & \multicolumn{3}{c}{Sentence Retrieval} & \multicolumn{3}{c}{Image Retrieval} \\
Method & R@1 & R@5 & R@10 & R@1 & R@5 & R@10 \\
\noalign{\smallskip}
\hline
\noalign{\smallskip}
VSE++ (fixed ResNet, 1 crop) \cite{faghri2017vse++} & 31.9 & - & 68.0 & 23.1 & - & 60.7  \\ 
Sum-Max t-i & 59.6 & 85.2 & 92.9 & 44.1 & 70.0 & 79.0 \\ 
Sum-Max i-t & 56.7 & 83.5 & 89.7 & 36.8 & 65.6 & 74.9 \\ 
SCO \cite{huang2017learning} (current state-of-the-art) & 55.5 & 82.0 & 89.3 & 41.1 & 70.5 & 80.1 \\ 
\hline
SCAN t-i AVG ($\lambda_1=9$)  & 61.8 & 87.5 & 93.7 & 45.8 & 74.4 & 83.0  \\ 
SCAN i-t AVG ($\lambda_1=10$) & 67.9 & 89.0 & 94.4 & 43.9 & 74.2 & 82.8  \\ 
\hline
\end{tabular}
\end{center}
\end{table}
\setlength{\tabcolsep}{1.4pt}

To begin with, we would like to incrementally validate our approach by revisiting a basic formulation of inferring the latent alignments between image regions and words without attention; {\em i.e.} the Sum-Max Text-Image proposed in \cite{karpathy2015deep} and its compliment, Sum-Max Image-Text (See Eqs. \eqref{eq:sum_max_ti} \eqref{eq:sum_max_it}). Our Sum-Max models adopt the same learning objectives with hard negatives sampling, bottom-up attention-based image representation, and sentence representation as SCAN. The only difference is that it simply aggregates the similarity scores of all possible pairs of image regions and words. The results and a comparison are presented in Table \ref{table:eff_sqattn}. VSE++ \cite{faghri2017vse++} matches whole images and full sentences on a single embedding vector. It uses pre-defined ResNet-152 trained on ImageNet \cite{deng2009imagenet} to extract one feature per image for training (single crop) and also leveraged hard negatives sampling, same as SCAN. 
Essentially, it represents the case without considering the latent correspondence but keeping other configurations similar to our Sum-Max models. The comparison between Sum-Max and VSE++ shows the effectiveness of inferring the latent alignments. 
With a better bottom-up attention model (compared to R-CNN in \cite{karpathy2015deep}), Sum-Max t-i even outperforms the current state-of-the-art. By comparing SCAN and Sum-Max models, we show that Stacked Cross Attention can further improve the performance significantly.

\begin{table}[t!]
\begin{center}
\caption{Effect of different SCAN configurations on Flickr30K. Results are reported in terms of Recall@$K$(R@$K$). i-t denotes Image-Text. SUM and MAX denote summation and max pooling instead of AVG/LSE at the pooling step, respectively.}
\label{table:ablation}
\begin{tabular}
{p{5.0cm}p{1.0cm}p{1.0cm}p{1.0cm}p{1.0cm}p{1.0cm}p{1.0cm}}
\hline\noalign{\smallskip}
 & \multicolumn{3}{c}{Sentence Retrieval} & \multicolumn{3}{c}{Image Retrieval} \\
Method & R@1 & R@5 & R@10 & R@1 & R@5 & R@10 \\
\noalign{\smallskip}
\hline
Baseline: SCAN i-t AVG & 67.9 & 89.0 & 94.4 & 43.9 & 74.2 & 82.8  \\ 
\hline
\noalign{\smallskip}
No hard negatives & 45.8 & 77.8 & 86.2 & 33.9 & 63.7 & 73.4  \\ 
Not normalize image embedding & 67.8 & 89.3 & 94.6 & 43.3 & 73.7 & 82.7 \\ 
SCAN i-t SUM  & 63.9 & 89.0 & 93.9 & 45.0 & 73.1 & 82.0 \\ 
SCAN i-t MAX  & 59.7 & 83.9 & 90.8 & 43.3 & 72.0 & 80.9  \\ 
One-directional GRU & 63.6 & 87.7 & 93.7 & 43.2 & 73.1 & 82.3  \\ 
\hline
\end{tabular}
\end{center}
\end{table}
\setlength{\tabcolsep}{1.4pt}

We further investigate in several different configurations with SCAN i-t AVG as our baseline model, and present the results in Table \ref{table:ablation}. Each experiment is performed with one alternation. It is observed that the gain we obtain from hard negatives in the triplet loss is very significant for our model, improving the model by 48.2\% in terms of sentence retrieval R@1. Not normalizing the image embedding (See Eq. \eqref{eq:cosine_sim}) changes the importance of image sample \cite{faghri2017vse++}, but SCAN is not significantly affected by this factor. Using summation (SUM) or maximum (MAX) instead of average or LogSumExp as the final pooling function yields weaker results. Finally, we find that using bi-directional GRU improves sentence retrieval R@1 by 4.3 and image retrieval R@1 by 0.7.

\begin{figure}[t!]
\centering
\includegraphics[width=\linewidth]{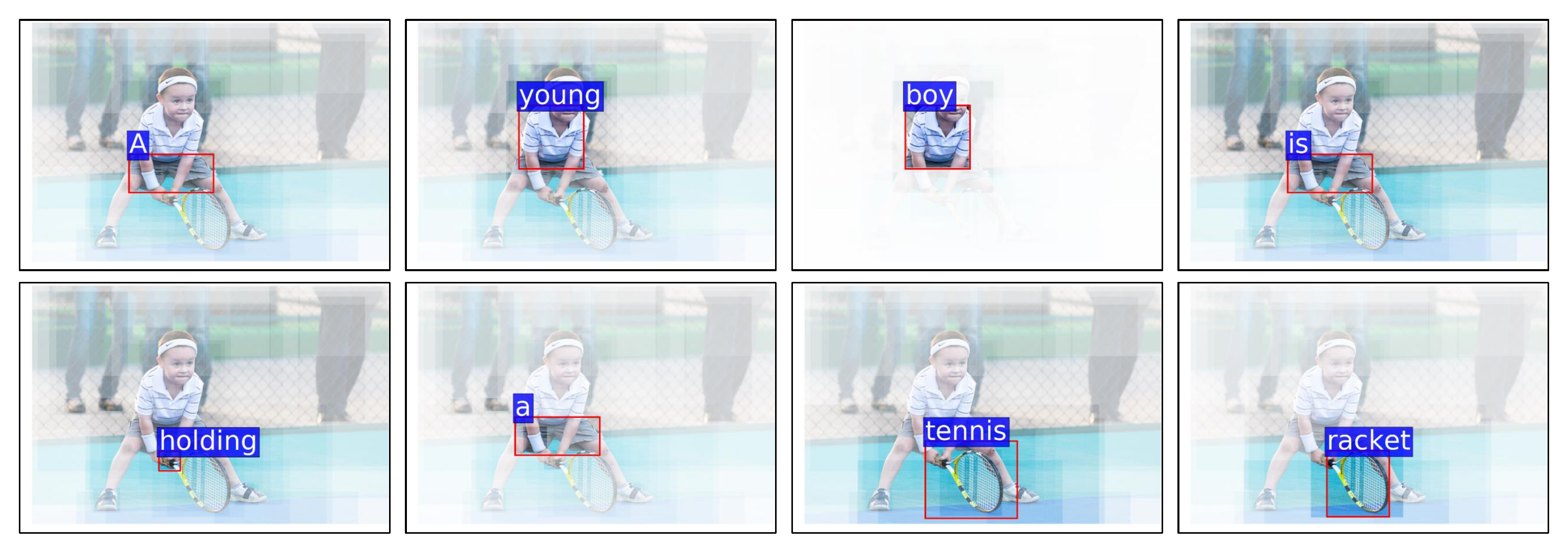}
\caption{Visualization of the attended image regions with respect to each word in the sentence description, outlining the region with the maximum attention weight in red. The regional brightness represents the attention strength, which considers the importance of both region and word estimated by our model. Our model generates interpretable focus shift and stresses on words like ``boy'' and ``tennis racket'', as well as the attributes (young) and actions (holding). (Best viewed in color)}
\label{fig:attn_vis}
\end{figure}

\section{Visualization and Analysis}

\subsection{Visualizing Attention}

By visualizing the attention component learned by the model, we are able to showcase the interpretablity of our model. In Figure \ref{fig:attn_vis}, we qualitatively present the attention changes predicted by our Text-Image model. For the selected image, we visualize the attention weights with respect to each word in the sentence description ``A young boy is holding a tennis racket.'' in different sub-figures. The regional brightness represents the attention weights which considers both importance of the region and the word corresponding to the sub-figure. We can observe that ``boy", ``holding", ``tennis" and ``racket" receive strong and focused attention on the relatively precise locations, while attention weights corresponding to ``a'' and ``is'' are weaker and less focused. This shows that our attention component learns interpretable alignments between image regions and words, and is able to generate reasonable focus shift and attention strength to weight regions and words by their importance while inferring image-text similarity.

\subsection{Image and Sentence Retrieval}
\begin{figure}[t!]
\centering
\includegraphics[width=\linewidth]{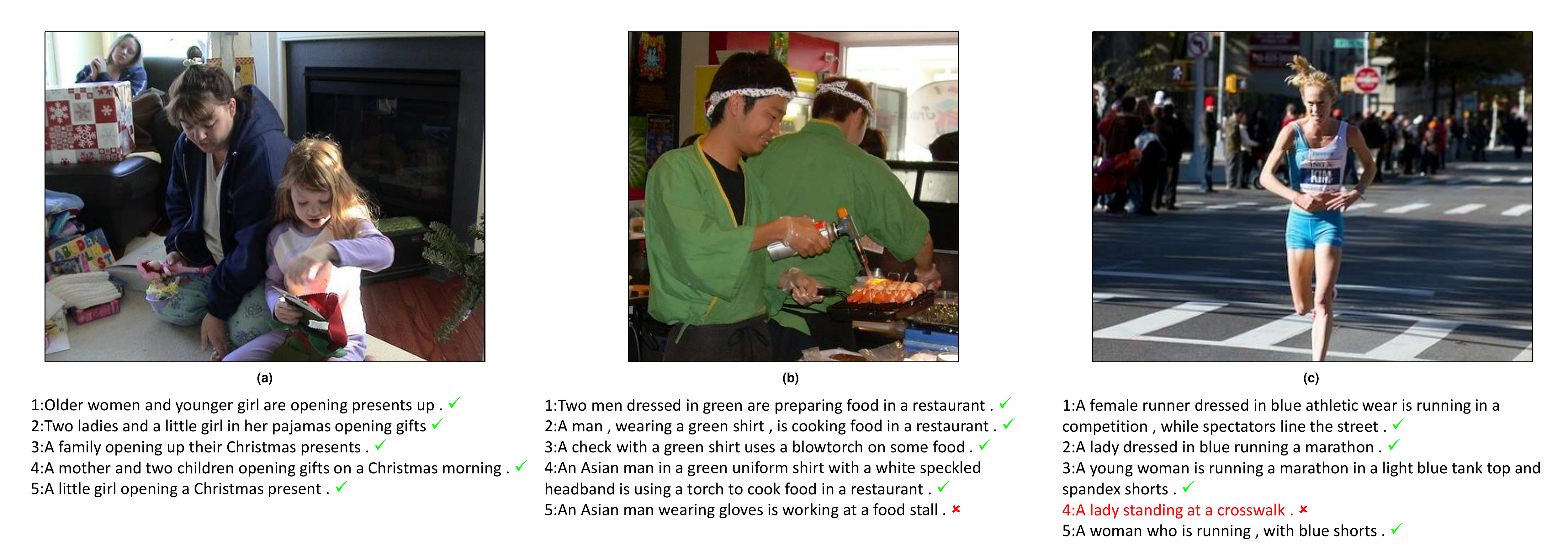}
\caption{Qualitative results of sentence retrieval given image queries on Flickr30K dataset. For each image query we show the top-5 ranked sentences. We observe that our Stacked Cross Attention model retrieves the correct results in the top ranked sentences even for image queries of complex and cluttered scenes. The model outputs some reasonable mismatches, {\em e.g.} (b.5). On the other hand, there are incorrect results such as (c.4), which is possibly due to a poor detection of action in static images. (Best viewed in color when zoomed in.)}
\label{fig:img_2_txt}
\end{figure}

\begin{figure}[t!]
\centering
\includegraphics[width=\linewidth]{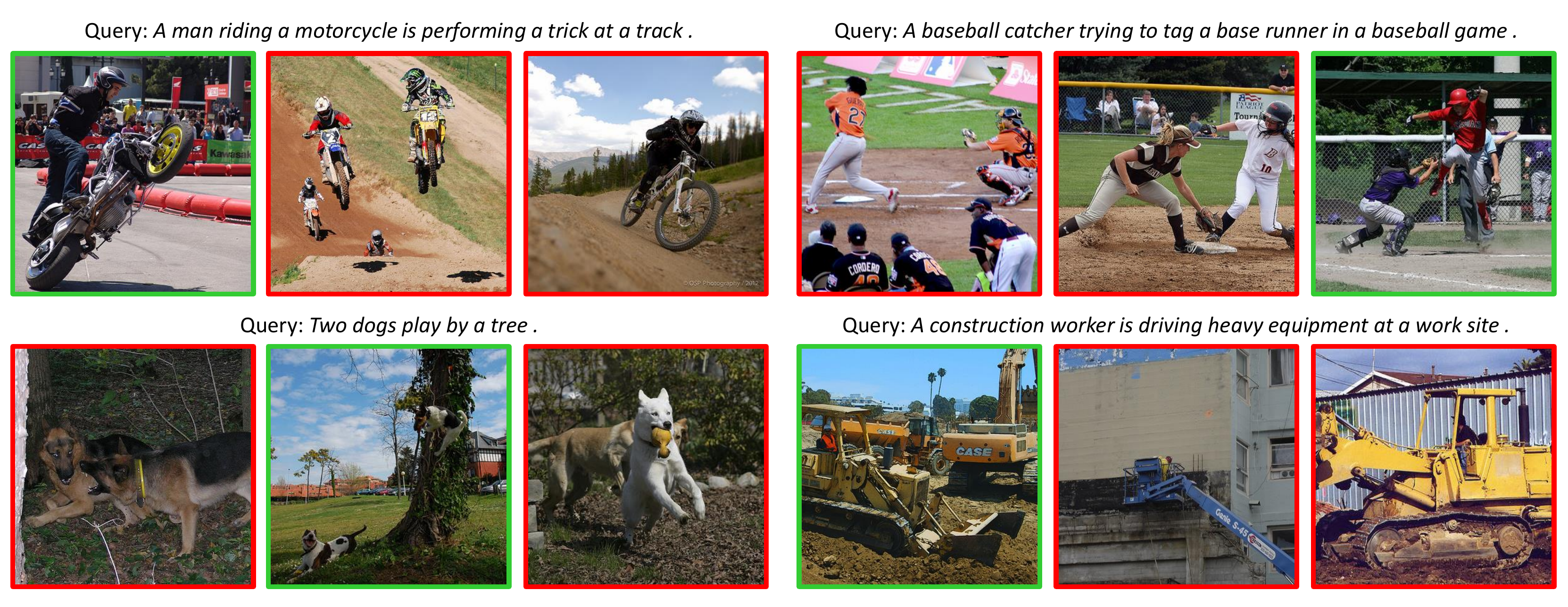}
\caption{Qualitative results of image retrieval given sentence queries on Flickr30K. For each sentence query, we show the top-3 ranked images, ranking from left to right. We outline the true matches in green boxes and false matches in red boxes. In the examples we show, our model retrieves the ground truth image in the top-3 list. Note that other results are also reasonable outputs. (Best viewed in color.)}
\label{fig:txt_2_im}
\end{figure}

Figure \ref{fig:img_2_txt} shows the qualitative results of sentence retrieval given image queries on Flickr30K. For each image query, we show the top-5 retrieved sentences ranked by the similarity scores predicted by our model. Figure \ref{fig:txt_2_im} illustrates the qualitative results of image retrieval given sentence queries on Flickr30K. Each sentence corresponds to a ground-truth image. For each sentence query we show the top-3 retrieved images, ranking from left to right. We outline the true matches in green and false matches in red.

\section{Conclusions}
We propose Stacked Cross Attention that gives the state-of-the-art performance on the Flickr30K and MS-COCO datasets in all measures. We carry out comprehensive ablation studies to verify that Stacked Cross Attention is essential to the performance of image-text matching, and revisit prior work to confirm the importance of inferring the latent correspondence between image regions and words. Furthermore, we show how the learned Stacked Cross Attention can be leveraged to give more interpretablity to such vision-language models. 

\noindent
\textbf{Acknowledgement.} The authors would like to thank Po-Sen Huang and Yokesh Kumar for helping the manuscript. We also thank Li Huang, Arun Sacheti, and Bing Multimedia team for supporting this work. Gang Hua is partly supported by National Natural Science Foundationof China under Grant 61629301.

\appendix
\section*{Appendix Overview}
This is a supplementary material for the main paper. Sec. \ref{sec:training_detail} presents the details of training the proposed Stacked Cross Attention Network. Sec. \ref{sec:roi_ablation} presents an ablation study on the number of Region of Interests (ROIs). Sec. \ref{sec:add_ex} presents additional qualitative examples of attended image regions, sentence retrieval for given image queries, and image retrieval for given sentence queries.

\section{Details of Training}
\label{sec:training_detail}
We use the Adam optimizer \cite{kingma2014adam} to train the models. For Flickr30K models, we train with a learning rate of 0.0002 for 15 epochs and then lower it to 0.00002 for another 15 epochs. For MS-COCO \cite{lin2014microsoft} models, we train with a learning rate of 0.0005 for 10 epochs and then lower the learning rate to 0.00005 for another 10 epochs. We set the margin of triplet loss, $\alpha$, to 0.2 (Eq. 14 of the main paper), mini-batch size to 128, and threshold of maximum gradient norm to 2.0 for gradient clipping. We also set the dimensionality of the GRU and the joint embedding space to
1024. The dimensionality of the word embeddings that are input to the GRU is set to 300. Using 1 Nvidia M40 GPU, training takes 11 hours on Flickr30K and 44 hours on MS-COCO. The source code has been made available at \url{https://github.com/kuanghuei/SCAN}.

\section{Details of Bottom-up Attention}
\label{sec:roi_ablation}
For visual bottom-up attention, we use Faster R-CNN model in conjunction with ResNet-101 pre-trained by Anderson et al. \cite{anderson2017bottom} to extract the Region of Interests (ROIs) for each image. The model is available at \url{https://github.com/peteanderson80/bottom-up-attention}. The Faster R-CNN implementation uses an intersection over union (IoU) threshold of 0.7 for region proposal suppression, and 0.3 for object class suppression. The top 36 ROIs with the highest class detection confidence scores are selected, following \cite{anderson2017bottom}. We extract features after average pooling, resulting in the final representation of 2048 dimensions.

\begin{table}[t!]
\begin{center}
\caption{Ablation on the effect of the number of ROIs, $k$. All experiments are done with the SCAN i-t AVG setting, where i-t denotes Image-Text and AVG denotes average pooling. Results are reported in terms of Recall@$K$(R@$K$). $\bar{T}_{sim}$ is the average running time (GPU) of computing similarity between an image and a sentence from encoded features using Stacked Cross Attention. $\bar{T}_{img}$ is the average time to encode image region features extracted from region detector for one image. $\bar{T}_{txt}$ is the average time to encode a sentence (not affected by $k$). $\bar{T}_{train}$ is the average training time for a mini-batch of 128 image-text pairs. } 
\label{table:abk}
\begin{tabular}
{p{0.6cm}p{0.9cm}p{0.9cm}p{0.9cm}p{0.9cm}p{0.9cm}p{0.9cm}p{1.2cm}p{1.2cm}p{1.2cm}p{1.4cm}}
\hline\noalign{\smallskip}
 & \multicolumn{3}{c}{Sentence Retrieval} & \multicolumn{3}{c}{Image Retrieval} \\
K & R@1 & R@5 & R@10 & R@1 & R@5 & R@10 & $\bar{T}_{sim}$ $\mu$s & $\bar{T}_{img}$ $\mu$s & $\bar{T}_{txt}$ $\mu$s & $\bar{T}_{train}$ $m$s\\
\noalign{\smallskip}
\hline
\noalign{\smallskip}
\multicolumn{10}{l}{Training/Inference with top $K$ regions} \\ 
12 & 63.1 & 85.0 & 91.2 & 39.2 & 69.7 & 79.2 & 6.48 & 2.71 & 53.74 & 745 \\ 
24 & 66.3 & 88.2 & 93.5 & 42.7 & 73.5 & 81.9 & 7.82 & 2.67 & 53.74 & 852 \\ 
36 & 67.9 & 89.0 & 94.4 & 43.9 & 74.2 & 82.8 & 9.09 & 2.47 & 53.74 & 989 \\ 
48 & 64.5 & 88.5 & 93.7 & 40.5 & 71.3 & 81.2 & 10.24 & 2.75 & 53.74 & 1112 \\ 
60 & 64.2 & 87.6 & 92.9 & 40.1 & 71.5 & 80.9 & 11.49 & 2.83 & 53.74 & 1287 \\ 
\hline
\multicolumn{10}{l}{Training with top 36 regions/Inference with $K$ regions} \\ 
12 & 60.9 & 84.7 & 92.4 & 42.0 & 70.8 & 79.9 & 6.48 & 2.71 & 53.74 & - \\ 
24 & 67.0 & 88.9 & 93.2 & 44.3 & 74.5 & 82.4 & 7.82 & 2.67 & 53.74 & - \\ 
\hline
\end{tabular}
\end{center}
\end{table}
\setlength{\tabcolsep}{1.4pt}

Table \ref{table:abk} presents an ablation study to show the effect of the number of ROIs, $k$, on model performance and running time, and verify the choice of $k=36$. When having the same $k$ for training and inference, $k=36$ yields the best results. We suspect the performance drops at 48 and 60 are caused by low-ranking regions that introduce noisy information. On the other hand, training and inference with smaller $k$ are faster (parallelized on GPU). Using 1 M40 GPU and $k=36$, one image retrieval query on Flickr30K test set (1000 images) takes around 10 ms, which is practical for re-ranking.

We also investigate the results of training with larger $k$ and testing with smaller $k$ in order to possibly save computation at inference time. However, we observe that using 12 or 24 regions at inference time for model trained with 36 regions results in similar drops comparing to using 12 or 24 regions at both training and inference time.

\section{Additional Examples}
\label{sec:add_ex}
\begin{figure}[b!]
\centering
\includegraphics[width=\textwidth]{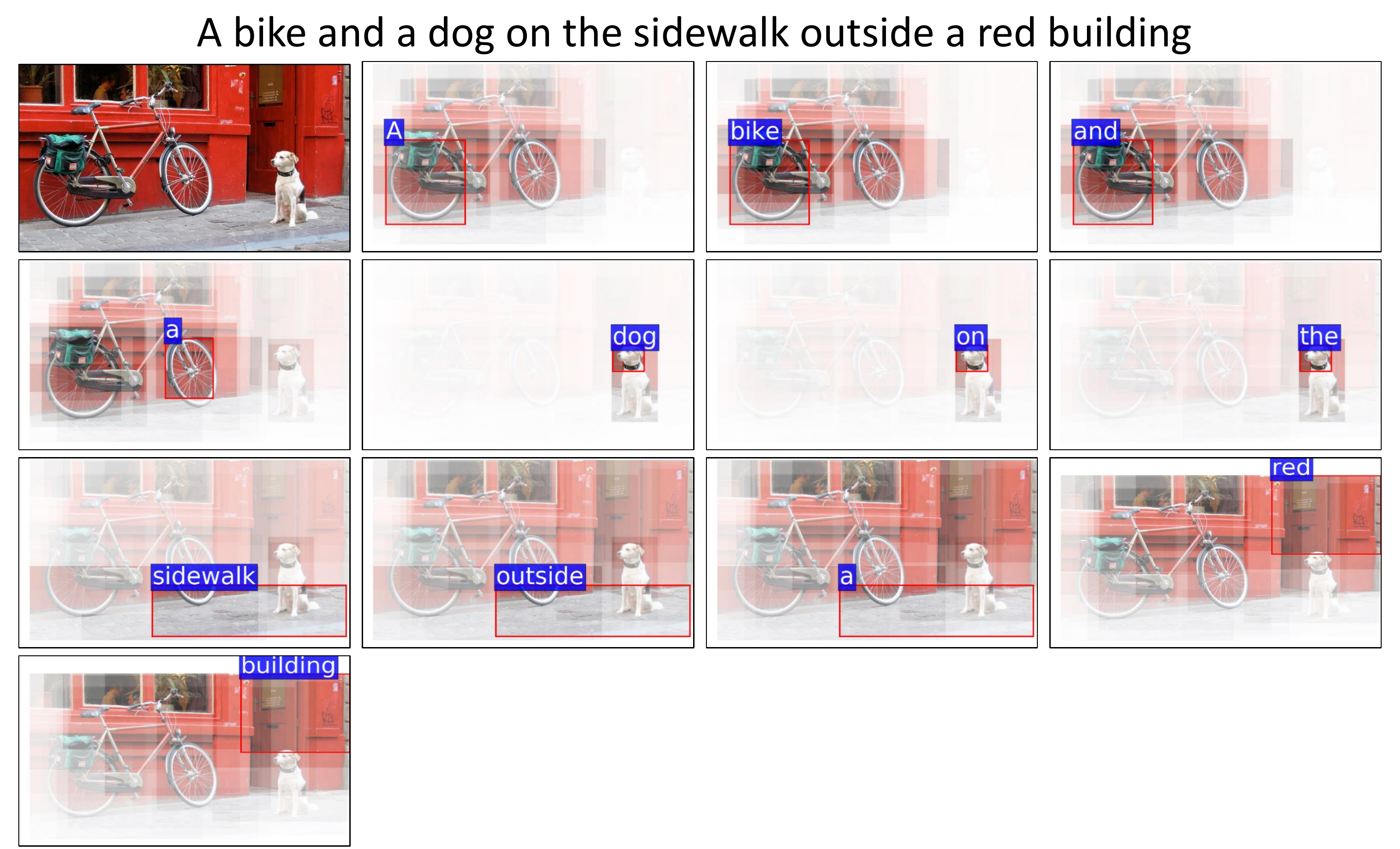}
\caption{An example of image-text matching showing attended image regions with respect to each word in the sentence. The brightness represents the attention strength, which considers the importance
of both regions and words estimated by our model. This example shows that our model can infer the alignments between words and the corresponding objects/stuff/attributes in the image (``bike'' and ``dog'' are objects; ``sidewalk'' and ``building'' are stuff; ``red'' is an attribute.)}
\label{fig:attnvis1}
\end{figure}

\begin{figure}
\centering
\includegraphics[width=\textwidth]{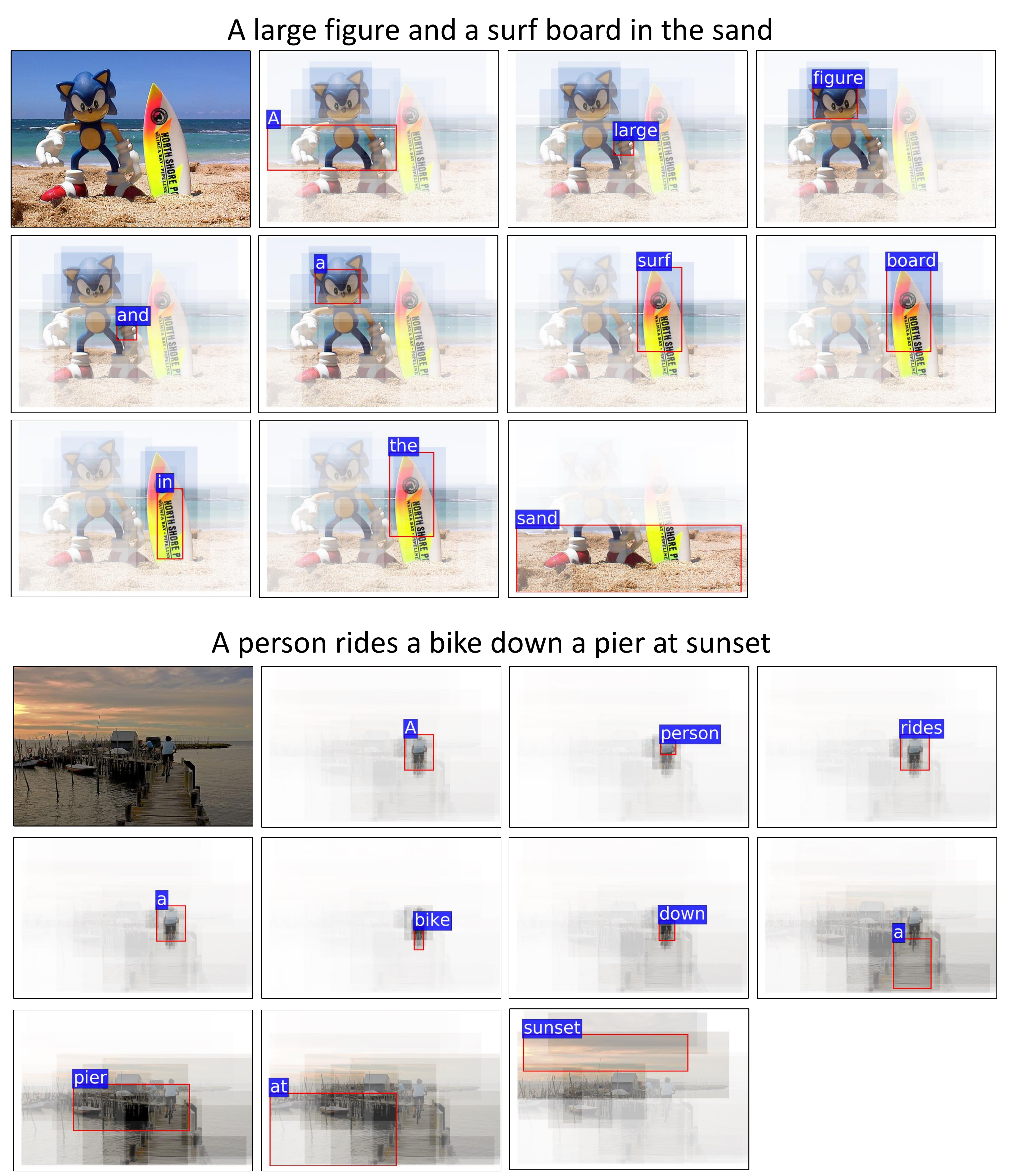}
\caption{Examples of image-text matching showing attended image regions with respect to each word. The brightness represents the attention strength, which considers the importance
of both regions and words estimated by our model. The two examples show that our model infers the alignments between words and the corresponding objects/actions/stuff in the images ({\em e.g.} for the bottom example, ``person'' and ``bike'' are objects; ``rides'' is an action; ``pier'' and ``sunset'' are stuff.)}
\label{fig:attnvis2}
\end{figure}

\begin{figure}
\centering
\includegraphics[width=\textwidth]{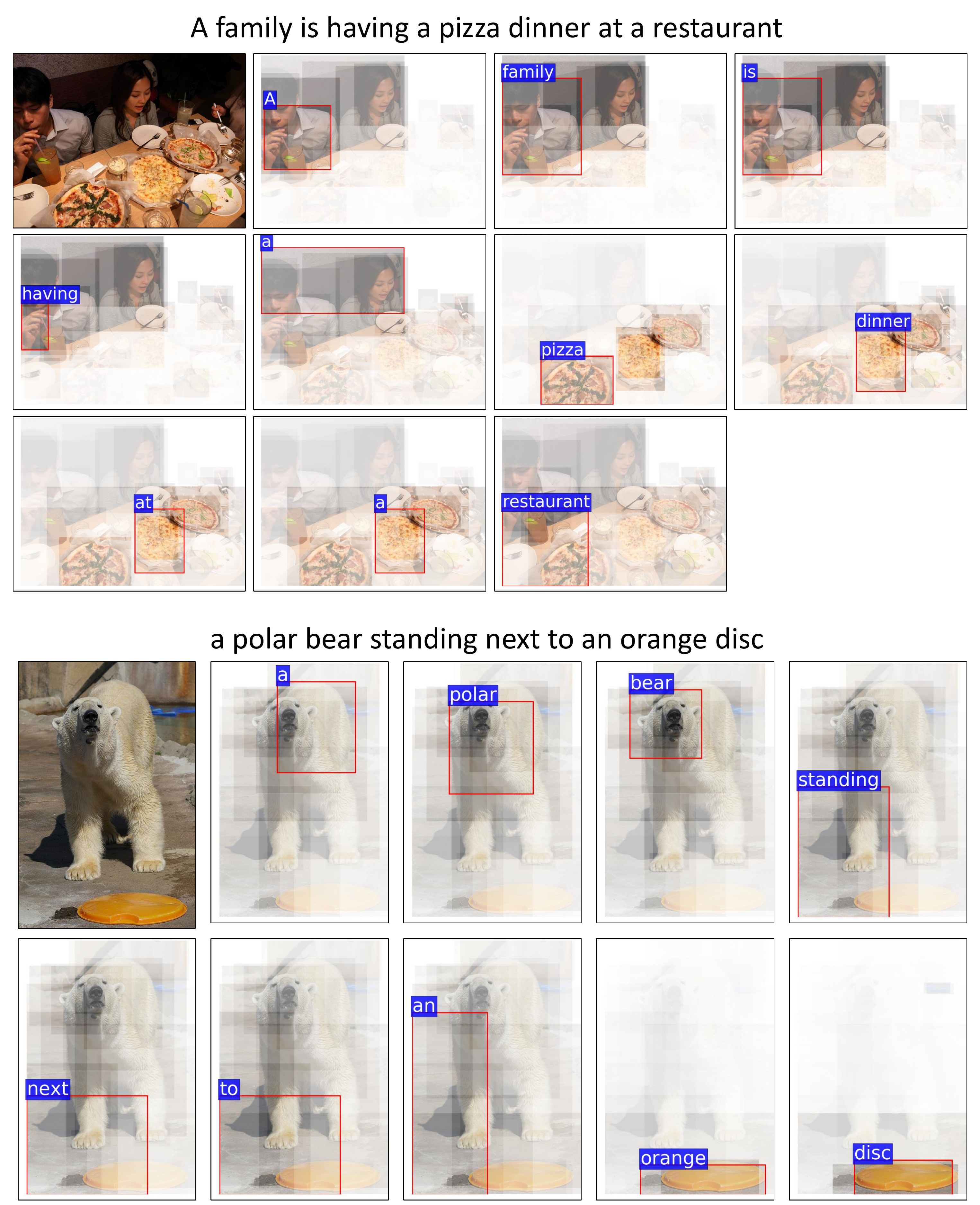}
\caption{Examples of image-text matching showing attended image regions with respect to each word. The brightness represents the attention strength, which considers the importance
of both regions and words estimated by our model. In the first image, we observe that focused attention is given to multiple objects when matching to words like ``family'' and ``pizza''. The bottom image suggests that attention is given to fine details such as the leg of the polar bear when matching to the word ``standing''.}
\label{fig:attnvis3}
\end{figure}

\begin{figure}[t!]
\centering
\includegraphics[width=\textwidth]{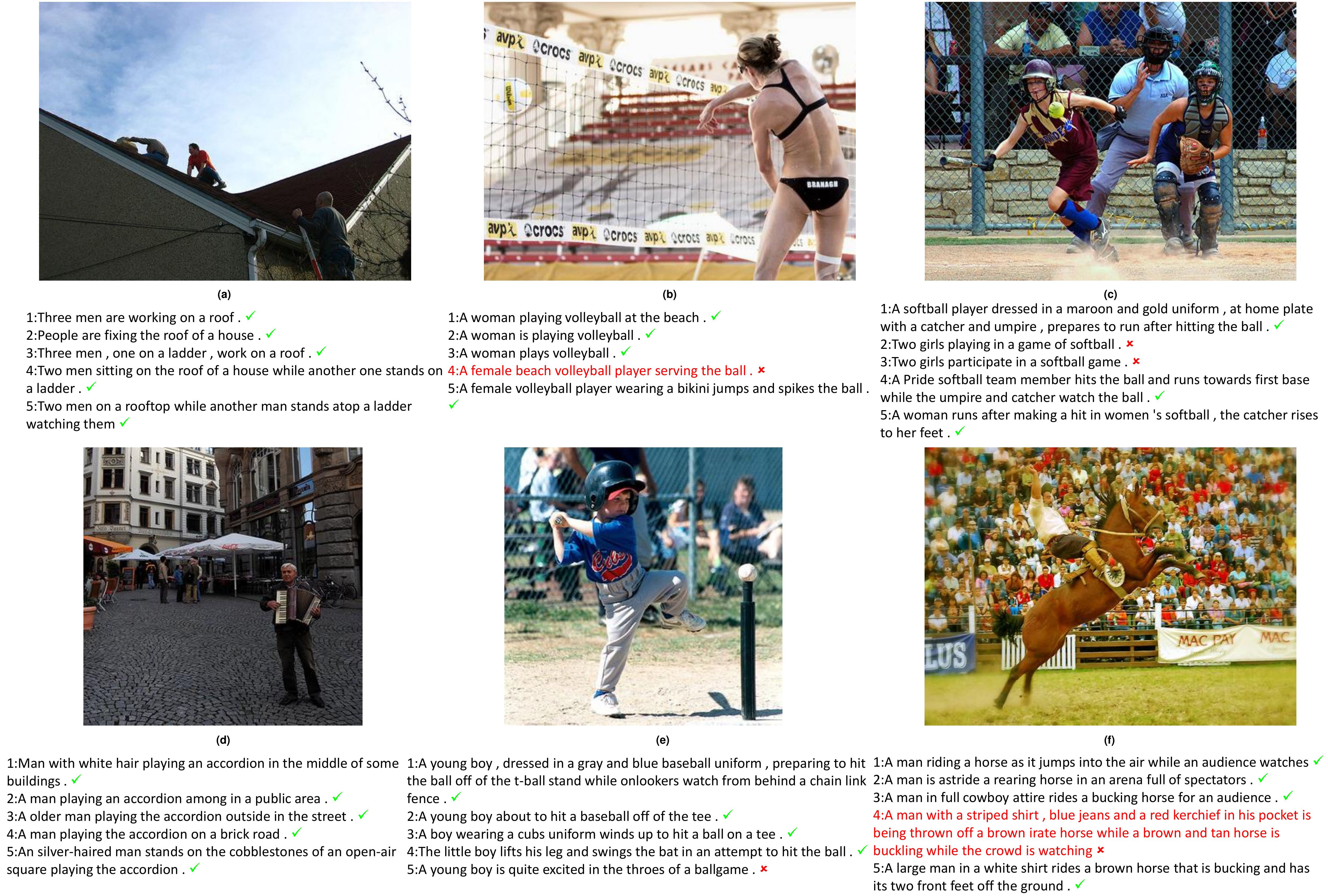}
\caption{Additional qualitative examples of text retrieval for given image queries on Flickr30K. Incorrect results are highlighted in red and marked with red x. Reasonable mismatches are in black but still marked with red x.}
\label{fig:f30k_txt_ret}
\end{figure}

\begin{figure}[b!]
\centering
\includegraphics[width=\textwidth]{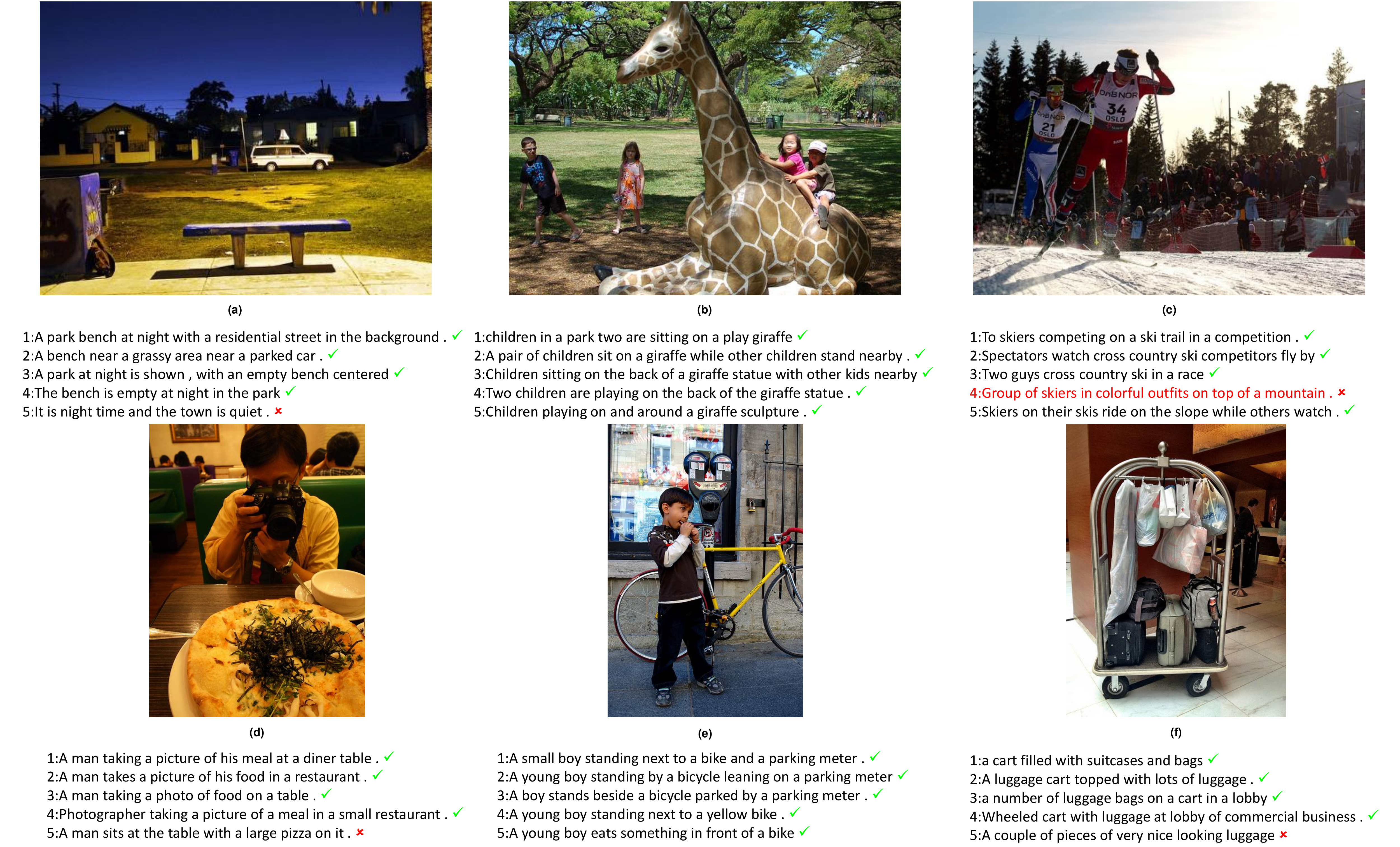}
\caption{Additional qualitative examples of text retrieval for given image queries on MS-COCO. Incorrect results are highlighted in red and marked with red x. Reasonable mismatches are in black but still marked with red x.} 
\label{fig:coco_txt_ret}
\end{figure}

\begin{figure}[b!]
\centering
\includegraphics[width=\textwidth]{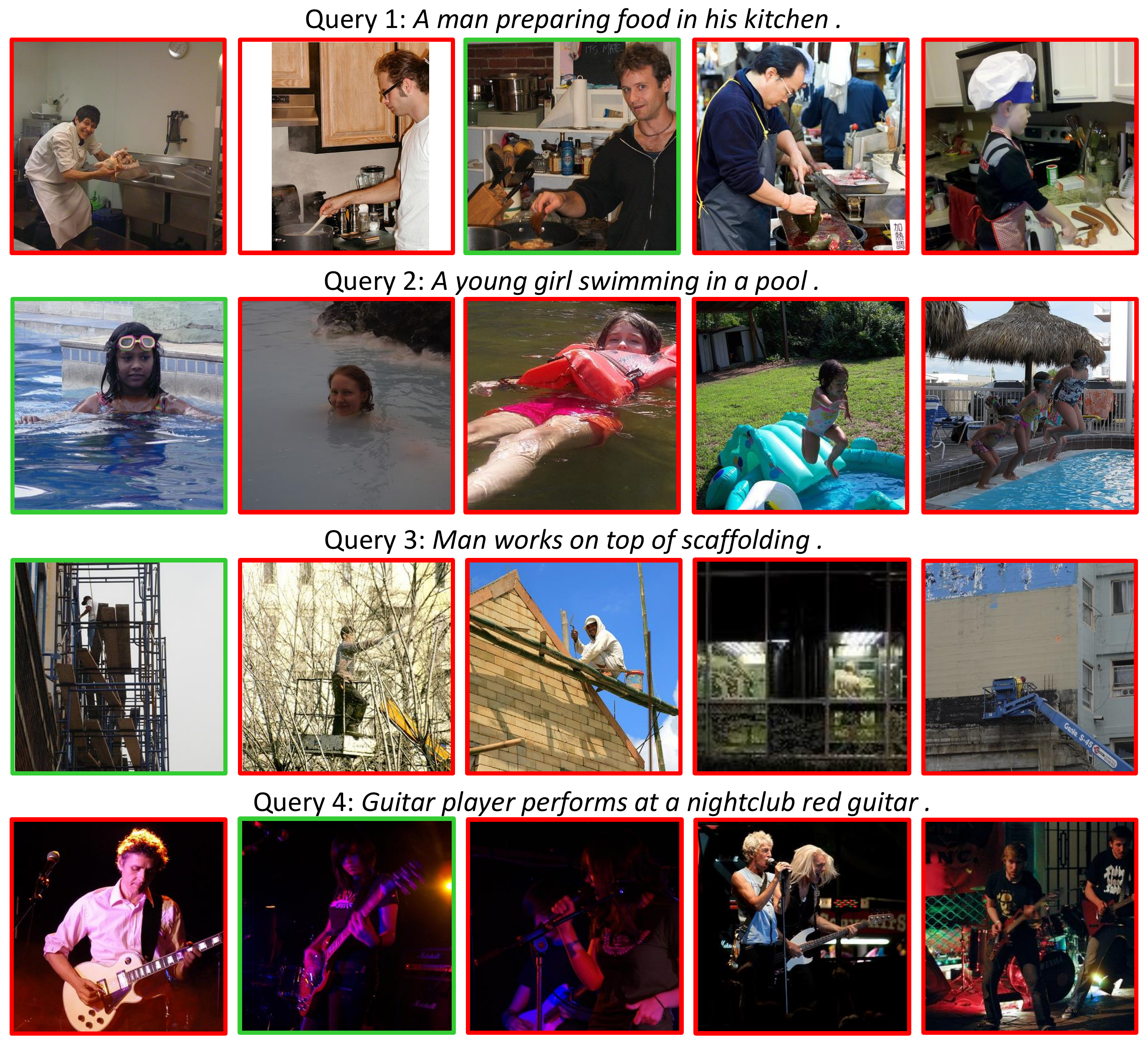}
\caption{Additional qualitative results of image retrieval for given sentence queries on Flickr30K. Each sentence description corresponds to one ground-truth image. For each sentence query, we show the top-5 ranked images, ranking from left to right. We outline the true matches in green and false matches in red. For query 1, our model ranks two reasonable mismatches before the ground-truth. The first output of query 4 is a failed case, where we observe that our attention component looks at the dark red light and the illuminated shirt for the word ``red''. Note that query 4 is grammatically incorrect. } 
\label{fig:f30k_im_ret}
\end{figure}

\begin{figure}[b!]
\centering
\includegraphics[width=\textwidth]{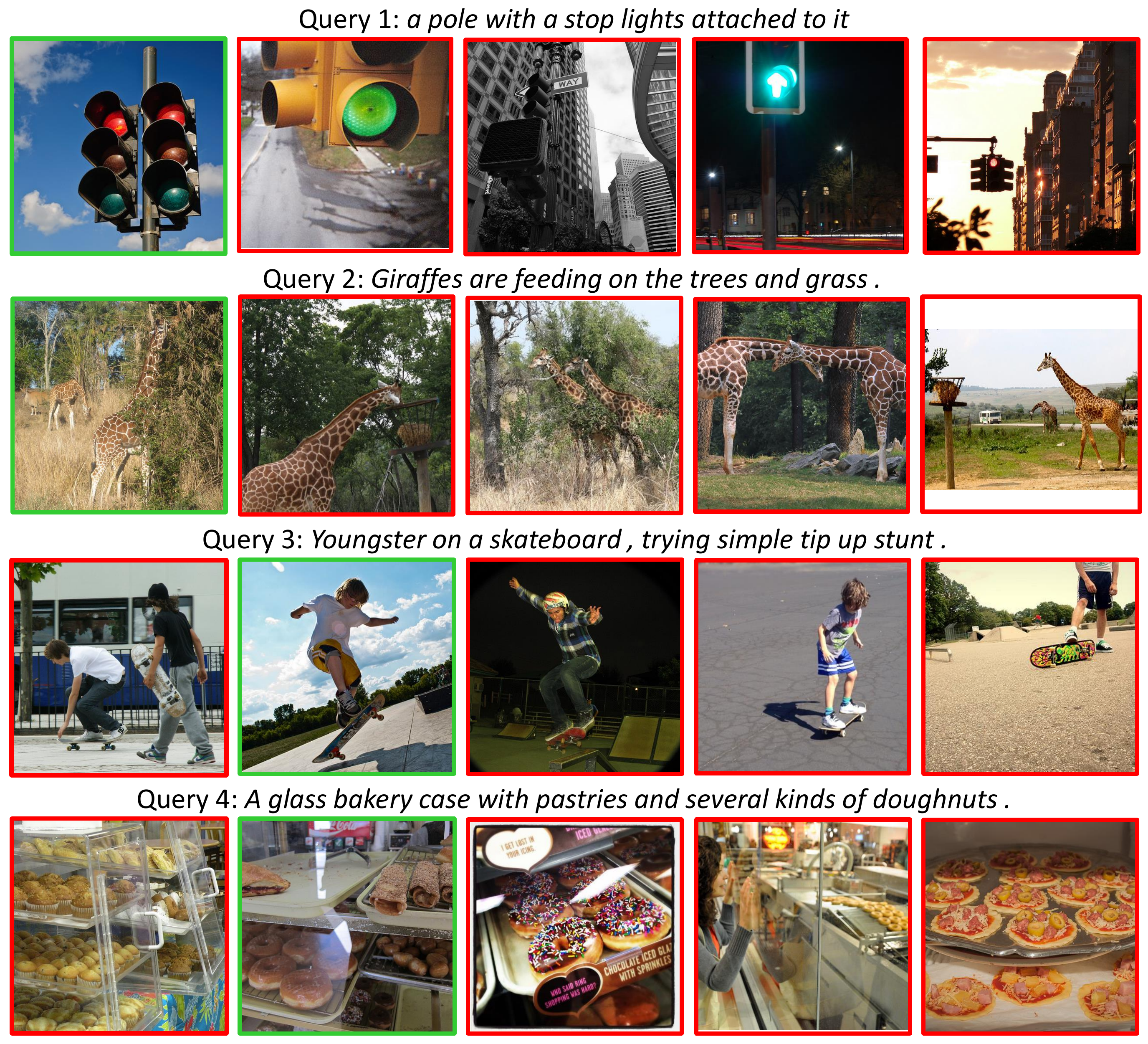}
\caption{Additional qualitative results of image retrieval for given sentence queries on MS-COCO. Each sentence description corresponds to one ground-truth image. For each sentence query, we show the top-5 ranked images, ranking from left to right. We outline the true matches in green and false matches in red. The first output of query 4 is a mismatch possibly caused by visual confusion. The bakery cases in the image are not glass but plastic. }
\label{fig:coco_im_ret}
\end{figure}

In this section, we present additional examples for qualitative analysis. We demonstrate additional examples of image-text matching (using a Text-Image Stacked Cross Attention Network) showing attended image regions in Figure \ref{fig:attnvis1}, Figure \ref{fig:attnvis2} and Figure \ref{fig:attnvis3}. Additional examples of sentence retrieval for given image queries on Flickr30K and MS-COCO can be found in Figure \ref{fig:f30k_txt_ret} and Figure \ref{fig:coco_txt_ret}, respectively. Furthermore, we show additional examples of image retrieval for given sentence queries on Flickr30K and MS-COCO in Figure \ref{fig:f30k_im_ret} and Figure \ref{fig:coco_im_ret}, respectively.

\clearpage

%
%
%
\bibliographystyle{splncs04}
\bibliography{mybibliography}

\begin{thebibliography}{10}
\providecommand{\url}[1]{\texttt{#1}}
\providecommand{\urlprefix}{URL }
\providecommand{\doi}[1]{https://doi.org/#1}

\bibitem{anderson2017bottom}
Anderson, P., He, X., Buehler, C., Teney, D., Johnson, M., Gould, S., Zhang,
  L.: Bottom-up and top-down attention for image captioning and {VQA}. In: CVPR
  (2018)

\bibitem{ba2016layer}
Ba, J.L., Kiros, J.R., Hinton, G.E.: Layer normalization. arXiv preprint
  arXiv:1607.06450  (2016)

\bibitem{bahdanau2014neural}
Bahdanau, D., Cho, K., Bengio, Y.: Neural machine translation by jointly
  learning to align and translate. In: ICLR (2015)

\bibitem{buschman2007top}
Buschman, T.J., Miller, E.K.: Top-down versus bottom-up control of attention in
  the prefrontal and posterior parietal cortices. Science  \textbf{315}(5820),
  1860--1862 (2007)

\bibitem{chorowski2015attention}
Chorowski, J.K., Bahdanau, D., Serdyuk, D., Cho, K., Bengio, Y.:
  Attention-based models for speech recognition. In: NIPS (2015)

\bibitem{corbetta2002control}
Corbetta, M., Shulman, G.L.: Control of goal-directed and stimulus-driven
  attention in the brain. Nature Reviews Neuroscience  \textbf{3}(3), ~201
  (2002)

\bibitem{deng2009imagenet}
Deng, J., Dong, W., Socher, R., Li, L.J., Li, K., Fei-Fei, L.: {ImageNet}: A
  large-scale hierarchical image database. In: CVPR (2009)

\bibitem{devlin2015language}
Devlin, J., Cheng, H., Fang, H., Gupta, S., Deng, L., He, X., Zweig, G.,
  Mitchell, M.: Language models for image captioning: The quirks and what
  works. In: ACL (2015)

\bibitem{eisenschtat2017linking}
Eisenschtat, A., Wolf, L.: Linking image and text with 2-way nets. In: CVPR
  (2017)

\bibitem{faghri2017vse++}
Faghri, F., Fleet, D.J., Kiros, J.R., Fidler, S.: {VSE++}: Improved
  visual-semantic embeddings. arXiv preprint arXiv:1707.05612  (2017)

\bibitem{fang2015captions}
Fang, H., Gupta, S., Iandola, F., Srivastava, R., Deng, L., Doll{\'a}r, P.,
  Gao, J., He, X., Mitchell, M., Platt, J., et~al.: From captions to visual
  concepts and back. In: CVPR (2015)

\bibitem{girshick2014rich}
Girshick, R., Donahue, J., Darrell, T., Malik, J.: Rich feature hierarchies for
  accurate object detection and semantic segmentation. In: CVPR (2014)

\bibitem{gu2017look}
Gu, J., Cai, J., Joty, S., Niu, L., Wang, G.: Look, imagine and match:
  Improving textual-visual cross-modal retrieval with generative models. In:
  CVPR (2018)

\bibitem{he2016deep}
He, K., Zhang, X., Ren, S., Sun, J.: Deep residual learning for image
  recognition. In: CVPR (2016)

\bibitem{he2008discriminative}
He, X., Deng, L., Chou, W.: Discriminative learning in sequential pattern
  recognition. IEEE Signal Processing Magazine  \textbf{25}(5) (2008)

\bibitem{huang2017instance}
Huang, Y., Wang, W., Wang, L.: Instance-aware image and sentence matching with
  selective multimodal {LSTM}. In: CVPR (2017)

\bibitem{huang2017learning}
Huang, Y., Wu, Q., Wang, L.: Learning semantic concepts and order for image and
  sentence matching. In: CVPR (2018)

\bibitem{juang1997minimum}
Juang, B.H., Hou, W., Lee, C.H.: Minimum classification error rate methods for
  speech recognition. IEEE Transactions on Speech and Audio processing
  \textbf{5}(3),  257--265 (1997)

\bibitem{karpathy2015deep}
Karpathy, A., Fei-Fei, L.: Deep visual-semantic alignments for generating image
  descriptions. In: CVPR (2015)

\bibitem{karpathy2014deep}
Karpathy, A., Joulin, A., Fei-Fei, L.: Deep fragment embeddings for
  bidirectional image sentence mapping. In: NIPS (2014)

\bibitem{katsuki2014bottom}
Katsuki, F., Constantinidis, C.: Bottom-up and top-down attention: Different
  processes and overlapping neural systems. The Neuroscientist  \textbf{20}(5),
   509--521 (2014)

\bibitem{kingma2014adam}
Kingma, D.P., Ba, J.: Adam: A method for stochastic optimization. In: ICLR
  (2015)

\bibitem{kiros2014unifying}
Kiros, R., Salakhutdinov, R., Zemel, R.S.: Unifying visual-semantic embeddings
  with multimodal neural language models. arXiv preprint arXiv:1411.2539
  (2014)

\bibitem{klein2015associating}
Klein, B., Lev, G., Sadeh, G., Wolf, L.: Associating neural word embeddings
  with deep image representations using fisher vectors. In: CVPR (2015)

\bibitem{krishna2017visual}
Krishna, R., Zhu, Y., Groth, O., Johnson, J., Hata, K., Kravitz, J., Chen, S.,
  Kalantidis, Y., Li, L.J., Shamma, D.A., et~al.: {Visual Genome}: Connecting
  language and vision using crowdsourced dense image annotations. International
  Journal of Computer Vision  \textbf{123}(1),  32--73 (2017)

\bibitem{kumar2016ask}
Kumar, A., Irsoy, O., Ondruska, P., Iyyer, M., Bradbury, J., Gulrajani, I.,
  Zhong, V., Paulus, R., Socher, R.: Ask me anything: Dynamic memory networks
  for natural language processing. In: ICML (2016)

\bibitem{lee2017cleannet}
Lee, K.H., He, X., Zhang, L., Yang, L.: {CleanNet}: Transfer learning for
  scalable image classifier training with label noise. In: CVPR (2018)

\bibitem{lev2016rnn}
Lev, G., Sadeh, G., Klein, B., Wolf, L.: {RNN} fisher vectors for action
  recognition and image annotation. In: ECCV (2016)

\bibitem{li2015hierarchical}
Li, J., Luong, M.T., Jurafsky, D.: A hierarchical neural autoencoder for
  paragraphs and documents. In: ACL (2015)

\bibitem{lin2014microsoft}
Lin, T.Y., Maire, M., Belongie, S., Hays, J., Perona, P., Ramanan, D.,
  Doll{\'a}r, P., Zitnick, C.L.: Microsoft {COCO}: Common objects in context.
  In: ECCV (2014)

\bibitem{luong2015effective}
Luong, M.T., Pham, H., Manning, C.D.: Effective approaches to attention-based
  neural machine translation. In: EMNLP (2015)

\bibitem{nam2016dual}
Nam, H., Ha, J.W., Kim, J.: Dual attention networks for multimodal reasoning
  and matching. In: CVPR (2017)

\bibitem{niu2017hierarchical}
Niu, Z., Zhou, M., Wang, L., Gao, X., Hua, G.: Hierarchical multimodal {LSTM}
  for dense visual-semantic embedding. In: ICCV (2017)

\bibitem{peng2017cm}
Peng, Y., Qi, J., Yuan, Y.: {CM-GANs}: Cross-modal generative adversarial
  networks for common representation learning. arXiv preprint arXiv:1710.05106
  (2017)

\bibitem{ren2015faster}
Ren, S., He, K., Girshick, R., Sun, J.: Faster {R-CNN}: Towards real-time
  object detection with region proposal networks. In: NIPS (2015)

\bibitem{rush2015neural}
Rush, A.M., Chopra, S., Weston, J.: A neural attention model for abstractive
  sentence summarization. In: EMNLP (2015)

\bibitem{schuster1997bidirectional}
Schuster, M., Paliwal, K.K.: Bidirectional recurrent neural networks. IEEE
  Transactions on Signal Processing  \textbf{45}(11),  2673--2681 (1997)

\bibitem{socher2014grounded}
Socher, R., Karpathy, A., Le, Q.V., Manning, C.D., Ng, A.Y.: Grounded
  compositional semantics for finding and describing images with sentences. In:
  ACL (2014)

\bibitem{vendrov2015order}
Vendrov, I., Kiros, R., Fidler, S., Urtasun, R.: Order-embeddings of images and
  language. In: ICLR (2016)

\bibitem{wang2016learning}
Wang, L., Li, Y., Lazebnik, S.: Learning deep structure-preserving image-text
  embeddings. In: CVPR (2016)

\bibitem{xu2015show}
Xu, K., Ba, J., Kiros, R., Cho, K., Courville, A., Salakhudinov, R., Zemel, R.,
  Bengio, Y.: Show, attend and tell: Neural image caption generation with
  visual attention. In: ICML (2015)

\bibitem{xu2017attngan}
Xu, T., Zhang, P., Huang, Q., Zhang, H., Gan, Z., Huang, X., He, X.: {AttnGAN}:
  Fine-grained text to image generation with attentional generative adversarial
  networks. In: CVPR (2018)

\bibitem{yang2016hierarchical}
Yang, Z., Yang, D., Dyer, C., He, X., Smola, A., Hovy, E.: Hierarchical
  attention networks for document classification. In: NAACL-HLT (2016)

\bibitem{young2014image}
Young, P., Lai, A., Hodosh, M., Hockenmaier, J.: From image descriptions to
  visual denotations: New similarity metrics for semantic inference over event
  descriptions. In: ACL (2014)

\bibitem{zheng2017dual}
Zheng, Z., Zheng, L., Garrett, M., Yang, Y., Shen, Y.D.: Dual-path
  convolutional image-text embedding. arXiv preprint arXiv:1711.05535  (2017)

\end{thebibliography}
%




\end{document}